\documentclass[sigconf]{acmart}

\AtBeginDocument{%
  \providecommand\BibTeX{{%
    \normalfont B\kern-0.5em{\scshape i\kern-0.25em b}\kern-0.8em\TeX}}}

\setcopyright{acmcopyright}
\copyrightyear{2020}
\acmYear{2020}
\acmDOI{}

\acmConference[Woodstock '20]{Woodstock '20}{2020}{Woodstock}
\acmBooktitle{Woodstock '20: ACM, 2020, Woodstock}
\acmPrice{15.00}
\acmISBN{978-1-4503-XXXX-X/18/06}
\usepackage{enumitem}
\usepackage{array}
\usepackage{multirow}
\usepackage{subfig}
\usepackage{xspace}
\usepackage{xcolor}
\usepackage{amsmath}
\usepackage{bm}
\newcommand{\mname}{\texttt{CovidCare}\xspace}

\usepackage{algorithm}  
\usepackage{algpseudocode}

\begin{document}

\title{\mname: Transferring Knowledge from Existing EMR to Emerging Epidemic for Interpretable Prognosis}




\author{
Liantao~Ma, Xinyu~Ma, Junyi~Gao, Chaohe~Zhang, Zhihao~Yu, Xianfeng~Jiao, Wenjie~Ruan, Yasha~Wang, Wen~Tang, and Jiangtao~Wang
}
\affiliation{ \institution{\textsuperscript{\rm 1}Key Laboratory of High Confidence Software Technologies, Ministry of Education, Beijing, China}}
\affiliation{ \institution{\textsuperscript{\rm 2}National Engineering Research Center of Software Engineering, Peking University, Beijing, China}}
\affiliation{ \institution{\textsuperscript{\rm 3}School of Electronics Engineering and Computer Science, Peking University, Beijing, China}}
\affiliation{ \institution{\textsuperscript{\rm 4}School of Computing and Communications, Lancaster University, UK}}
\affiliation{ \institution{\textsuperscript{\rm 5}Division of Nephrology, Peking University Third Hospital, Beijing, China }}
\email{malt@pku.edu.cn}

\begin{abstract}

Due to the characteristics of COVID-19, the epidemic develops rapidly and overwhelms health service systems worldwide.
Many patients suffer from systemic life-threatening problems and need to be carefully monitored in ICUs.
Thus the intelligent prognosis is in an urgent need to assist physicians to take an early intervention, prevent the adverse outcome, and optimize the medical resource allocation.
However, in the early stage of the epidemic outbreak, the data available for analysis is limited due to the lack of effective diagnostic mechanisms, rarity of the cases, and privacy concerns.
In this paper, we propose a deep-learning-based approach, \mname, which leverages the existing electronic medical records to enhance the prognosis for inpatients with emerging infectious diseases.
It learns to embed the COVID-19-related medical features based on massive existing EMR data via transfer learning.
The transferred parameters are further trained to imitate the teacher model's representation behavior based on knowledge distillation, which embeds the health status more comprehensively in the source dataset.
We conduct the length of stay prediction experiments for patients on a real-world COVID-19 dataset.
The experiment results indicate that our proposed model consistently outperforms the comparative baseline methods.
\mname also reveals that, {\em 1)} \textit{hs-cTnI}, \textit{hs-CRP} and \textit{Platelet Counts} are the most fatal biomarkers, whose abnormal values usually indicate emergency adverse outcome.
{\em 2)} Normal values of \textit{$\gamma$-GT}, \textit{AP} and \textit{eGFR} indicate the overall improvement of health. 
The medical findings extracted by \mname are empirically confirmed by human experts and medical literatures.

\end{abstract}




\maketitle

\section{Introduction}









\begin{figure}[]
  \centering
  \includegraphics[width=0.95\columnwidth]{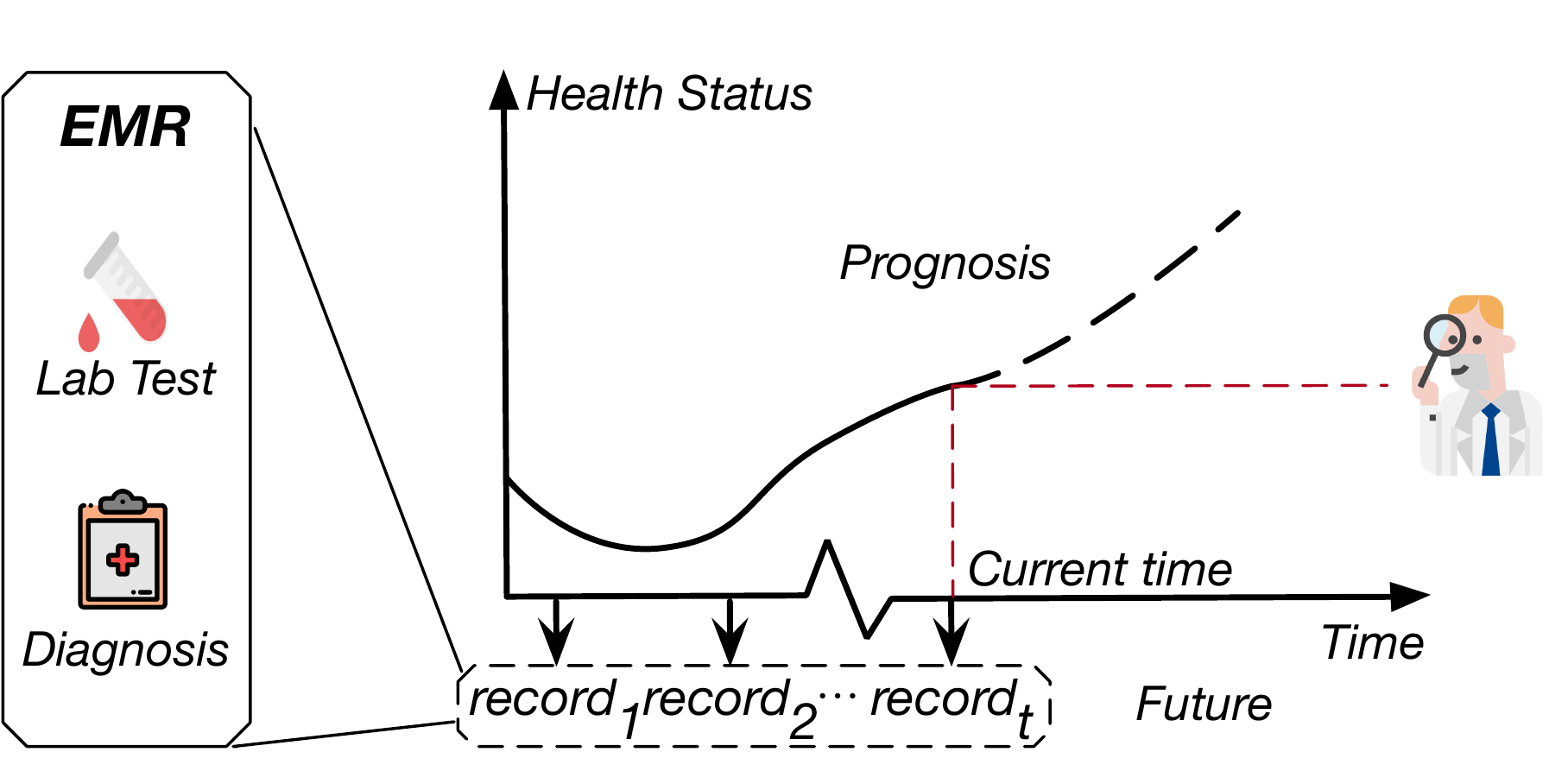}
  \caption{Performing health evaluation is crucial for early clinical intervention and medical resource management.}
  \label{fig:introduction}
\end{figure}

The whole world is now facing the unprecedented crisis brought by COVID-19.
The exponential growth of COVID-19 patients has brought massive pressure on the health systems tragically, such as overwhelming the national health service and exhausting the intensive care units (ICUs). 
It is crucially essential to personalize prognosis for the individual patient by considering her/his specific health condition to enable a timely and early medical intervention, as shown in Figure~\ref{fig:introduction}. 
The accurate prediction of the remaining length-of-stay for inpatients is critical for scheduling and optimizing limited hospital resources~\cite{harutyunyan2019multitask}. 


However, for newly-emerged infectious diseases (e.g., COVID-19, SARS) and rare diseases, the prognosis performed by human physicians may not meet the clinical demand, especially in rural areas and developing countries.
Moreover, the precise risk prediction requires a high level of clinical expertise and experience~\cite{pedersen1990prospective}. 
%
%
However, the accumulation of clinical experience is time-consuming and difficult at the early outbreak of the new emerging infectious disease (EID).
Thus it is difficult for human physicians to comprehensively evaluate the health of patients and accurately identify the key factors, especially in a situation where the deterioration of some EIDs in the early stage is usually not evident~\cite{lauer2020incubation}. So during treating COVID-19, it is not so rare that physicians omit the ominous signs and miss the chance of early intervention, especially when the clinical resources are insufficient.





As a result, 
intelligent prognosis is in an urgent need against EID and rare diseases.
It not only can assist physicians to perform early diagnosis, select personalized treatments, and prevent adverse outcomes, but also optimize the allocation of medical resources and reduce the medical cost~\cite{weissman2020locally}.
Recently, many deep learning-based models have been developed to enable intelligent prognosis by analyzing electronic medical records (EMRs), including mortality prediction~\cite{ma2020adacare,ma2020concare}, disease diagnosis prediction~\cite{lee2018diagnosis}, and patient phenotype identification~\cite{baytas2017patient}.
To enrich the feature extraction and health status representation, most existing research works utilize sophisticated modules to extract health status representations that require a large amount of labeled training data.
However, the quantity of labeled clinical data available for prognosis may not be ideal in practice in the early stage of the EID outbreak~\cite{huang2020clinical}, due to the following reasons. {\em 1)} The precise diagnostic mechanism has not been established in the early outbreak.
Before introducing the nucleic acid detection mechanism, it is difficult to confirm whether a patient is really infected with COVID-19.
For example, there are only 41 patients diagnosed with COVID-19 due to the lack of valid testing methods at the early outbreak in Wuhan,~\cite{huang2020clinical}.
{\em 2)} The disease is still progressing for patients. The collection of enough outcomes needs to take a long time.
{\em 3)} There are serious privacy concerns about electronic medical records, so the data-sharing mechanism for EID across multiple hospitals worldwide usually cannot be established timely.
Therefore, the scarce labeled data will decrease the performance of deep learning models due to the potential over-fitting.


Recently some researchers try to exploit additional information to deal with the scarcity of clinical data.
On one hand, some researches encode ontology resources and structured relationships among medical codes (e.g., diagnosis of diabetes) in the network to enhance the representation learning.
For example, GRAM~\cite{choi2017gram} and KAME~\cite{ma2018kame} introduce the external well-organized ontology information (e.g., International Classification of Diseases Codes) to represent the medical concept as a combination of its ancestors in the ontology via an attention mechanism.
However, for the new EIDs like COVID-19, such relationship and ontology information are also difficult to acquire.
On the other hand, some researchers try to make full use of the existing time series data through transfer learning.
For instance, Doctor AI~\cite{Choi2015Doctor} and Gupta~\cite{gupta2018transfer} train deep models at one hospital and transfer them to another hospital.
However, these methods can only be adapted for the same tasks with similar clinical features between the source and target dataset.
TimeNet \cite{gupta2018transfer} has been trained on different non-clinical time series datasets via an RNN autoencoder in an unsupervised manner to extract generic features for patient phenotyping. 
However, the extracted general-purpose features may not be suitable for a specific clinical task, leading to the underperformance of the model.

Therefore, for the prognosis of EIDs with limited data, such a research challenge remains:
{\em How to make full use of the existing EMR data to learn the robust health status representation, when tackling tasks with different clinical feature sets?}

In this paper, we propose a novel healthcare predictive approach, \mname, based on transfer learning from existing EMR data (i.e., source dataset) to the new dataset (i.e., target dataset) with knowledge distillation. 
To improve the compatibility across source dataset and target dataset with different feature sets, 
\mname evaluates the health status of patients mainly from the perspective of clinical features rather than visits. The time series of each feature is embedded separately by GRUs. When training on the target dataset, the shared features of both datasets are specifically encoded by a pre-trained GRU.
The model with private features trained on the source dataset is treated as a teacher network to guide the embedding behavior of the shared features. Doing so is able to further explore and leverage the information stored in the source dataset.
Finally, feature-wise attention is deployed to abstract the biomarkers and adaptively identify the critical features for patients in diverse health conditions. In summary, \mname contributes to the community from the following aspects:

\begin{itemize}[leftmargin=*]

\item We propose a transfer-learning-based medical feature embedding approach, \mname, to perform clinical prediction for EIDs with limited data .
Multi-channel architecture is developed to improve the compatibility across source and target datasets with different feature sets. By jointly optimizing the prediction loss and similarity loss, the student model with shared features is learned to imitate the teacher model's encoding behavior with full features on the source dataset.
We further use the feature re-calibration module to provide the interpretability, which explicitly enhances high-risk features.

\item We conduct length-of-stay prediction experiments for inpatients with COVID-19. 
The results show that \mname significantly and consistently outperforms the baseline approaches for all evaluation metrics.


\item \mname can extract valuable medical findings for COVID-19: 
\begin{itemize}
    \item \textit{Hypersensitive Cardiac Troponin}, \textit{Hypersensitive C-Reactive Protein} and \textit{Platelet Counts} are the most fatal biomarkers, whose abnormal values usually indicate emergency adverse outcome.
    \item Normal values of \textit{$\gamma$-Glutamyl Transpeptidase}, \textit{Alkaline Phosphatase} and \textit{estimated Glomerular Filtration Rate} indicate the overall improvement of health.
\end{itemize}
We invite medical practitioners to evaluate the extracted medical knowledge and prognosis cases.
Human experts positively confirmed the clinical significance in the aspects of early prediction, key biomarkers extraction, and clinical resources management. 

\item Beyond COVID-19, we also conduct mortality risk prediction for outpatients with end-stage renal diseases (ESRD) to verify the applicability of \mname to other diseases with limited EMR.
The extensive experiments demonstrate that \mname can significantly benefit the prognosis for future pandemics and rare diseases.

\end{itemize}

\section{Related Work}

\subsection{Prognosis for COVID-19}
Outbreaks of the COVID-19 epidemic have been causing worldwide health concerns and was officially declared a pandemic by the World Health Organization (WHO) on March 11, 2020. Although the ultimate impact of COVID-19 is uncertain, it has significantly overwhelmed health care infrastructure. All emerging viral pandemics can place extraordinary and sustained demands on public health and health systems and providers of essential community services~\cite{us2017pandemic}. Limited health-care resource availability will increase the chance of being infected while waiting for treatment and also the mortality rate~\cite{ji2020potential}. This eventually leads to an increase in the severity of the pandemic. The rapidly growing imbalance between supply and demand for medical resources in many countries presents an inherent normative question: How can we make early and accurate risk prediction to allocate medical resources effectively during a pandemic?

Many COVID-related researches focus on the severity of disease rather than the clinical outcome of mortality~\cite{emami2020prevalence,wang2020comorbid,fu2020clinical}. These studies answer key clinical questions on COVID-19 evolution and outcomes, as well as potential risk factors leading to hospital and ICU admission. However, they cannot make individualized risk predictions for patients. Recently, Li et al.~\cite{yan2020interpretable} use machine learning-based methods such as decision tree to make risk prediction for COVID-19 patients. However, as discussed above, many challenges, such as data scarcity and model interpretability, have not been adequately addressed. To optimize patient care and appropriately deploy health care resources during this pandemic, effective and reliable early risk prediction is still an essential and urgent problem.

\subsection{Deep-Learning-Based EMR Analysis}

With the prevalence of electronic healthcare information systems in various healthcare institutions, a large amount of Electronic Medical Records (EMR) have been accumulated over time\cite{lee2017big,reddy2015healthcare}.
EMR is a type of multivariate time series data that records patients' visits in hospitals (e.g., diagnoses, lab tests, as shown in Figure~\ref{fig:dynamic_data}).
This provides essential healthcare information for the data-driven clinical status prediction.
Deep learning-based models have shown the capability to perform
mortality prediction
\cite{esteban2016predicting,suresh2018learning,heo2018uncertainty,gao2020stagenet},
patients subtyping
\cite{baytas2017patient},
and
diagnosis prediction
\cite{lee2018diagnosis,choi2016retain,ma2018health,qiao2018pairwise,pham2016deepcare,baytas2017patient}.
For most of the researches, extracting advanced clinical features and learning the compressed representation of the sparse EMR data are fundamental procedures of clinical healthcare prediction.

EMR is longitudinally complex \cite{zheng2017resolving, choi2018mime}.
Extracting the advanced clinical representation would introduce more parameters into the model, making the model more complex and hard to train.
For EIDs and some rare diseases, the quantity of labeled data is much less, which can not support a model to be trained thoroughly.
In order to deal with this issue, some researches try to introduce additional information about the data.

On one hand, GRAM~\cite{choi2017gram} and KAME~\cite{ma2018kame} incorporate the external medical information (e.g., ontologies of the medical codes), which makes the model to be trained more sufficiently. 
They exploit medical knowledge in the whole prediction process by using a given medical ontology (i.e., knowledge graph), such as the International Classification of Diseases (ICD), to learn the representations of medical codes and obtain the embeddings of medical codes' ancestors.
MIME \cite{choi2018mime} learns the multi-level embedding of data according to the knowledge about the inherent EMR structure (e.g., the multi-level relationship among medical codes).
However, such external structured information and the extra knowledge about the data are often not easy to be accessed or used in the clinical practice for EIDs.
Ontology information is usually designed to handle the medical codes. 
It is not suitable for dealing with numerical lab tests, which also are essential clinical features to capture health status.

On the other hand, some researchers try to explore the existing EMR data.
Choi~\cite{Choi2015Doctor} empirically confirms that RNN models possess great potential for transfer learning across different medical institutions.
Gupta~\cite{gupta2018transfer} trains a deep RNN to identify several patient phenotypes on time series from MIMIC-III database, and then uses the features extracted using that RNN to build classifiers for identifying previously unseen phenotypes.
However, these methods can only be utilized for the same tasks with the same clinical feature sets between source and target datasets. 
TimeNet~\cite{gupta2018using} is pre-trained on non-medical time series in an unsupervised manner and further utilized to extract features for clinical prediction.
Nevertheless, the extracted general-purpose features may not be suitable for exploring the specific clinical task, thus leading to limited performance.

\section{Problem Formulation}

\begin{table}[]
    \centering
       \caption{Notations used in \mname }
    \label{tab:notations}
    \resizebox{\columnwidth}{!}{
    \begin{tabular}{l|l}
        \hline
        Notation & Definition \\
         \hline
         $y_{T, tar}$ & Groundtruth of LOS prediction at $T$-th admission on target dataset \\
         $\hat{y}_{T, tar}$ & Prediction result at $T$-th admission on source dataset \\
         $y_{T, src}$ & Groundtruth of prediction at $T$-th admission on source dataset \\
         $\hat{y}_{T, src}$ & Prediction result at $T$-th admission on source dataset \\
         \hline
         $\mathcal{R}_{src}$ & The whole source dataset \\
         $\mathcal{R}_{tar}$ & The whole target dataset \\
         $\tilde{\mathcal{R}}_{tar}$ & Target dataset (only included shared features with source dataset)\\
         \hline
         $\bm{r}_{i}$ & A time series record of the $i$-th medical feature \\
         $\bm{f}_{i}$ & Embedding of the $i$-th medical feature \\
         $\bm{f}_{i}^{*}$ & Embedding of the $i$-th medical feature after self-attention \\
         $\bm{s}$ & Overall representation of patient \\
         \hline
         $X_{tea}$ & Model/Embeddings/Parameters used in Teacher model \\
         $X_{stu}$ & Model/Embeddings/Parameters used in Student model \\
         $X_{tar}$ & Model/Embeddings/Parameters used in Target model \\
         \hline
         
    \end{tabular}}
\end{table}


Many patients suffering from COVID-19 face severe life threats and need careful health monitoring and medical treatment in ICU.
Typically, some biomarkers, such as \textit{Hypersensitive C-Reactive Protein}, are recorded through the treatment trajectories, and further have been taken into consideration for the prognosis. 
Accurate prediction of health status can help with assessing the severity of illness; and determining the value of novel treatments, interventions, and health care policies \cite{purushotham2017benchmark}.
Besides, due to the characteristics of COVID-19, large numbers of sick people appear for treatment during peak illness periods.
Clinics and hospitals are overwhelmed.
Predicting remaining time spent in ICU (i.e., length of stay) for admission is also vital for scheduling and hospital resource management. 

Below we define the data and task studied in this work and provide the list of notations used in \mname in Table \ref{tab:notations}.



\begin{figure}[]
\centering
\includegraphics[width=0.95\columnwidth]{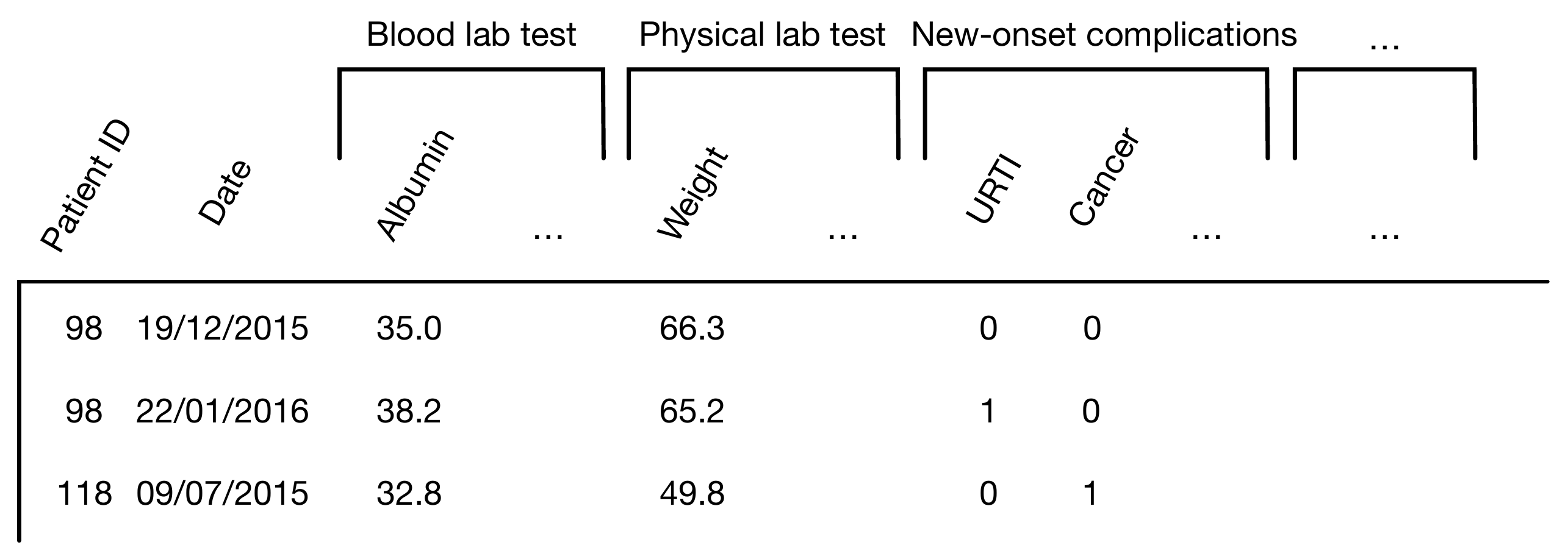}
\caption{Medical features. The physician conducts the necessary lab tests for the patient at each admission.}
\label{fig:dynamic_data}
\end{figure}

\textbf{Definition: Electronic Medical Records (EMR)}
Electronic Medical Records (EMR) data are routinely collected patient observations from hospitals through the clinical admissions, including discrete time-series data (e.g., medication, diagnosis) and continuous multivariate data (e.g., vital signs, laboratory measurements), as shown in Figure~\ref{fig:dynamic_data}. 
The admissions generating $N$ features such as different lab test results denoted as $\bm{r}_{i}\in\mathbb{R}^{T}$ $(i = 1, 2,$ $\cdots$ $, N)$. 
Each medical feature contains $T$ timesteps, as shown in Figure \ref{fig:dynamic_data}.
As a result, such a clinical sequence can be formulated as a “longitudinal patient matrix” $record$, where one dimension represents medical features and the other denotes admission timestamps~\cite{lee2017big}.


\begin{figure*}[]
\centering
\includegraphics[scale = 0.35]{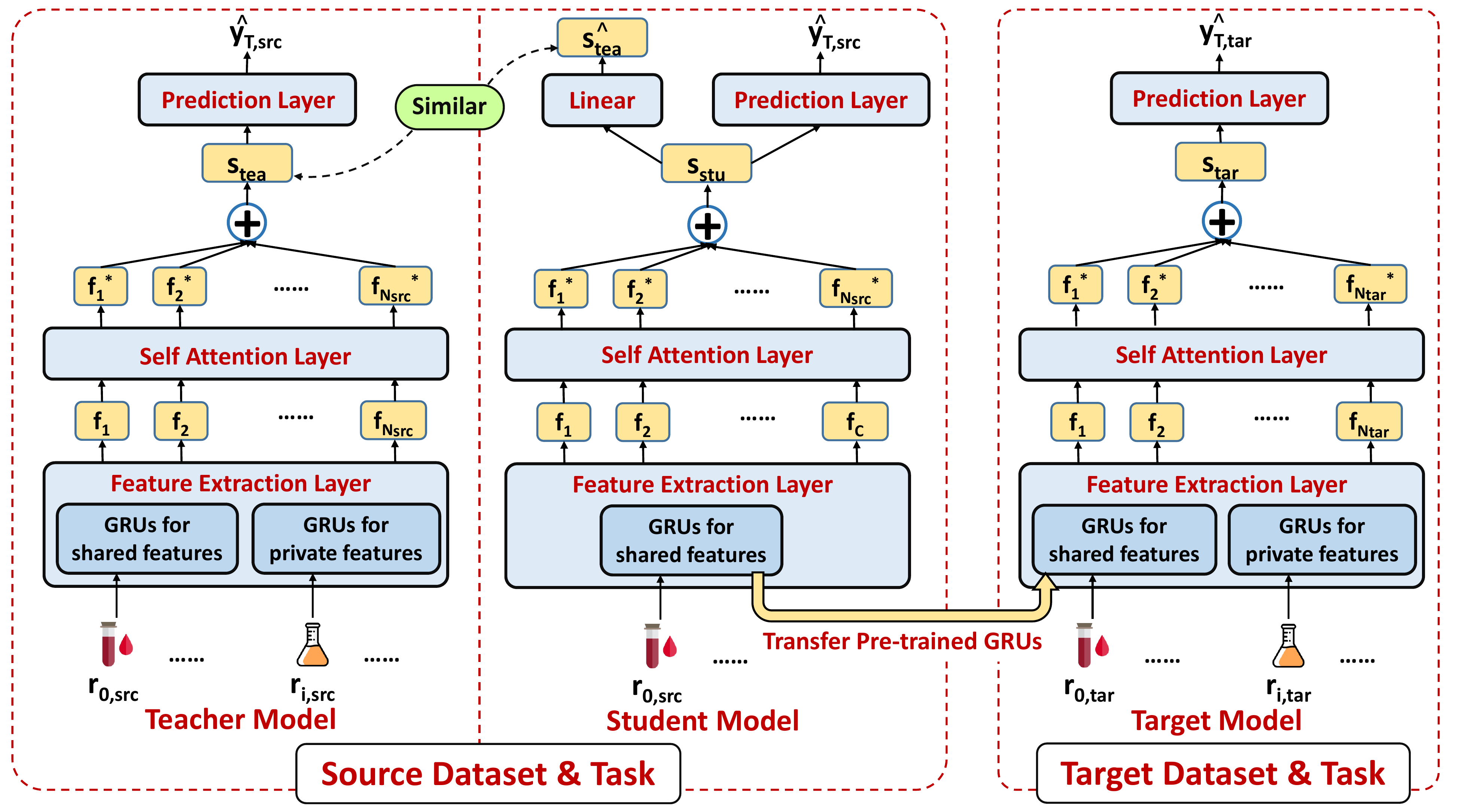}
\caption{The \mname Framework}
\label{fig:framework}
\end{figure*}

\textbf{Problem: Length of stay prediction}
The prediction problem in this paper can be formulated as given $N_r$ historic EMR data of a patient, i.e., $(\bm{r}_{1}$, $\cdots$ , $\bm{r}_{N})$, how to predict the patient's remaining time spent in ICU, $\hat{y}$ (i.e., length of stay).
We frame length of stay (LOS) prediction as a classification problem with 12 classes (discharging in 1/2/3/5/10/10+ days, suffering adverse outcome in 1/2/3/5/10/10+ days).

\section{Methodology}

\subsection{Overview}

Figure~\ref{fig:framework} shows the framework of the proposed \mname, which comprises of the following procedures.
The whole model training process is shown in the Appendix, Algorithm~\ref{alg:full_model}.

\begin{itemize}[leftmargin=*]

\item Multivariate time series with all features are fed into the healthcare representation learning module as a teacher model to build the comprehensive embedding in the source dataset. 

\item The student model in the source dataset learns to embed the proper health status based on features that are shared with the target dataset, by imitating the teacher model's embedding behavior.

\item The learned parameters of feature embedding are transferred to the healthcare representation learning model for the target dataset, and further fine-tuned to perform the task-specific prediction.

\end{itemize}

\subsection{Healthcare Representation Learning}

In this subsection, we will introduce the patient health status embedding module based on ConCare~\cite{ma2020concare}, which is a healthcare context representation learning method. There are three layers designed in this module, which consist of the feature extracting layer, self-attention layer, and prediction layer.

We utilized the multichannel GRU mechanism in the feature extracting layer to capture different patterns of each medical feature individually, which is also designed to improve the scalability and compatibility of the feature-specific model transfer. 
Specifically, we apply $N$ different GRUs to embed the $N$ medical features. Each feature $i$ can be described as a time series $\bm{r_i} = (r_{i1}, r_{i2}, \cdots, r_{iT})$, and will be fed into the corresponding GRU$_{i}$ to generate feature embedding $f_{i}$: $\bm{f_{i}} = \mathrm{GRU}_{i}(r_{i1}, r_{i2}, \cdots, r_{iT})$. The feature embedding matrix $\mathbf{F} = (\bm{f_{1}},\bm{f_{2}},\cdots,\bm{f_{N}}) ^ {\mathsf{T}}$.

In the self-attention layer, we employ the multi-head self-attention mechanism to obtain information from the health context and better understand the correlations between medical features.
This mechanism makes each feature adaptively interact with all other features, and combine information from the related ones according to self-attention weights.
Mathematically, the self-attention weight matrix of head $i$: 
\begin{equation}
    \mathcal{A}_{i} = \mathrm{Softmax} (\frac{\mathbf{Q}_{i} \mathbf{K}_{i}^{\mathsf{T}}}{\sqrt{d_{k}}})
\end{equation}
where $\mathbf{Q}_{i} = \mathbf{F} \cdot \mathbf{W}{^{\mathbf{Q}}_{i}} , \mathbf{K}_{i} = \mathbf{F} \cdot \mathbf{W}{^{\mathbf{K}}_{i}} $, and $d_{k}$ is the size of the row vector of matrix $\mathbf{K}_{i}$. 
And the result of feature interaction in head $i$:
\begin{equation}
    head_{i} = \mathcal{A}_{i} \mathbf{V}_{i}
\end{equation}
where $\mathbf{V}_{i} = \mathbf{F} \cdot \mathbf{W}{^{\mathbf{V}}_{i}}$.
And finally, the embedding matrix $\mathbf{F}^{*}$ after feature interaction: 
\begin{equation}
    \mathbf{F}^{*} = (\bm{f}_{1}^{*},...,\bm{f}_{N}^{*})^ {\mathsf{T}}
 = (head_{0} \oplus head_{1} \oplus ... \oplus  head_{m}) \mathbf{W}^{O}
\end{equation}
 
We also utilize the attention mechanism to integrate embeddings of all features $\bm{f}_{i}^{*}$ into an overall representation of patient $\bm{s}$, and interpret the importance of medical features at the same time.
Eventually, in the prediction layer, we apply a full-connection layer to conduct corresponding prediction tasks, and we select cross-entropy as the loss term $\mathcal{L}_{pred}$.
\begin{equation}
\mathcal{L}_{pred} = \mathrm{CE}(\hat{y}_T, y_T) = - \frac{1}{B} \sum_{b=1}^{B} [ y{_T^b}\log(\hat{y}{_T^b}) + (1-y{_T^b})\log(1 - \hat{y}{_T^b})]
\end{equation}

\subsection{Knowledge Distillation}

Firstly, we will introduce the feature-specific transfer learning mechanism.
As we mentioned before, a small data volume of the dataset may restrict a deep learning model's training performance. If our target task only consists of relatively few data items, it is necessary to make use of the existing EMR dataset with a bigger data volume to help our model training. Because the information and pattern extracted from a bigger dataset are always more stable and general, which is also useful to our target model.

Based on the patient health status embedding module introduced above, we conduct feature-specific transfer learning on feature extraction layer, since this layer mainly captures the general pattern of medical features, which is independent of patient cohorts and prediction tasks. 
Concretely, we transfer GRUs of shared features from the source model to the target model, so that we can make up for the shortcomings of small data volume by transfer obtained knowledge from a bigger dataset. 

However, we have not sufficiently extracted information from our source dataset since some private features remain unused, and it is obvious that unused features in the source dataset can also provide sufficient significant information.
In other words, with a complete source dataset, we can capture correlations between features more sufficiently, and thus generating a more comprehensive representation of patients.

Therefore, we proposed a knowledge distillation method to construct a more comprehensive transfer source model. We divide the source model into two parts, teacher model and student model. 
The student model is trained on the source dataset with only shared features ($\tilde{\mathcal{R}}_{src}$) and will be transferred to the target model. While the teacher model is trained on the complete dataset with all features ($\mathcal{R}_{src}$), but only auxiliary to student model and will not be transferred.
Specifically, 'Knowledge distillation' exactly means we hope the student model could imitate the behavior of the teacher model to learn a more comprehensive representation of patients, just like the teacher model does. 
We design an additional loss term for this, which encourages the student model to restore the representation learned by the teacher model.

In detail, we first train the teacher model to generate representation $s_{tea}$ for every patient with loss term $\mathcal{L}_{tea} = \mathcal{L}_{pred}$. We then train the student model, where the representation $s_{stu}$ should perform two functions, to predict corresponding task labels, and to imitate $s_{tea}$ by a linear layer as much as possible. We use KL-divergence to calculate the similarity of the two representations.
\begin{equation}
    \hat{s}_{tea} = s_{stu} \cdot \mathbf{W}_{stu}
\end{equation}
\begin{equation}
    \mathcal{L}_{emb} = D_{KL} (\mathrm{Softmax}(\hat{s}_{tea}) || \mathrm{Softmax}(s_{tea}) ) 
\end{equation}
\begin{equation}
      D_{KL}(P||Q) = \sum_{i} P_i \log(\frac{P_i}{Q_i})
\end{equation}
The loss of student model ($\mathcal{L}_{stu}$) is described as the sum of two parts accordingly, $\mathcal{L}_{stu} = \mathcal{L}_{pred} + \mathcal{L}_{emb}$.
And finally we tranfer GRUs from student model to target model, and fine tune the target model with target dataset($\mathcal{R}_{tar}$) using loss term $\mathcal{L}_{tar} = \mathcal{L}_{pred}$ . 


\section{Experiment}

We conduct the experiment that leveraging data from PhysioNet Source Dataset~\cite{reyna2019early} to enhance the LOS prediction for COVID-19~\cite{yan2020interpretable}.

\begin{table}
  \caption{Statistics of the Datasets }
  \label{tab:dataset}
  \begin{tabular}{lcc}
    \toprule
    Dataset &PhysioNet& COVID-19\\
    \midrule
    \# of patients & 40,336 & 375\\
    \# of admissions & 1,552,210 &  6,120\\
    Avg. \# of admissions per patient & 38.48 & 16.32\\
    Max. \# of admissions per patient & 336 &  59\\
    Min. \# of admissions per patient & 8 &  1\\
    \# of features & 33 &  74\\
    \# of adverse outcomes & 2,932 &   174 \\
    \% of adverse outcomes & 7.26\% &  46.40\% \\
  \bottomrule
\end{tabular}
\end{table}

\subsection{Data Description}

\subsubsection{COVID-19 target Dataset}

We take the COVID-19 dataset \cite{yan2020interpretable} as the target dataset and perform the LOS prediction.
The medical information of all patients collected between 10 January and 18 February 2020 was used for model development.
The average age of the patients was 58.83 years, and 59.7\% were male.
Of the 375 cases included in the subsequent analysis, 201 recovered from COVID-19 and were discharged from the hospital, while 174 died.
Statistics of source dataset and target dataset are listed in the Appendix.
Statistics of the LOS are listed in Table~\ref{tab:statlos}.
Without loss of generality, we perform the length of stay prediction for patients at 10th admission in this paper.
The distribution of days to the outcome for admissions are shown in Figure~\ref{fig:days2outcome}.
Medical features recorded in COVID-19 target dataset are listed in Table~\ref{tab:feature}.

\begin{table*}[]
\small 
\centering
\caption{Length of Stay Prediction Performance}
\label{tab:results}
\begin{tabular}{cccccc} 
\toprule
 &  & \multicolumn{4}{c}{Length of Stay Prediction Performance on COVID-19 Dataset} \\
Methods & Transfer  & AUPRC & AUROC-Macro & AUROC-Micro & min(Se, P+)  \\ 
\midrule
GRU & $\times$  & 0.2146\,(0.0343) & 0.6636\,(0.0464) & 0.7325\,(0.0362) & 0.2608\,(0.0435) \\

MC-GRU & $\times$  & 0.2603\,(0.0331) & 0.7702\,(0.0334) & 0.8044\,(0.0247) & 0.3038\,(0.0276) \\

ConCare & $\times$  & 0.3046\,(0.0312) & 0.7792\,(0.0214) & 0.8199\,(0.0186) & 0.3483\,(0.0114)\\
TimeNet & $\surd$  & 0.2908\,(0.0325) & 0.7752\,(0.0299) & 0.8093\,(0.0229) & 0.3354\,(0.0356) \\

\hline

MC-GRU$_{t}$ & $\surd$  & 0.2946\,(0.0331) & 0.7761\,(0.0300) & 0.8155\,(0.0189) & 0.3413\,(0.0308) \\

\mname$_{stu}$ & $\surd$  & 0.2989\,(0.0342) & 0.7768\,(0.0160) & 0.8146\,(0.0130) & 0.3236\,(0.0231)\\

\mname & $\surd$  & \textbf{0.3252}\,(0.0457) & \textbf{0.7859}\,(0.0202) & \textbf{0.8245}\,(0.0230) & \textbf{0.3538}\,(0.0246)\\

\bottomrule
\end{tabular}
\end{table*}

\subsubsection{PhysioNet Source Dataset}

We take the PhysioNet Dataset~\cite{reyna2019early} as the source dataset and pre-train the medical feature embedding based on the Sepsis prediction. 
This dataset is sourced from ICU patients in two separate U.S. hospital systems.
These data were collected over the past decade with approval from the appropriate Institutional Review Boards. They are labeled by Sepsis-3 clinical criteria.
The cleaned dataset consists of 40,336 patients and consists of a combination of hourly vital sign summaries (e.g., heart rate, systolic blood pressure), laboratory values (e.g., Chloride, Glucose). 
In particular, the data contained 33 clinical variables: 8 vital sign variables and 25 laboratory variables.
The statistics of the datasets are presented in Table~\ref{tab:dataset}.
Medical features recorded in the PhysioNet source dataset are listed in Table~\ref{tab:feature}.









\subsection{Experimental Setup}

\subsubsection{Evaluation Preparation}

Due to the limited amount of data, 5-fold cross-validation is employed on the prediction task. We do not separate independent test data.
We assess the performance of multi-classification using the area under the receiver operating characteristic curve (AUROC-Micro/Macro), area under the precision-recall curve (AUPRC), the minimum of precision and sensitivity Min(Se, P+).
Note that $Micro$ calculates metrics globally by considering each element of the label indicator matrix as a label. $Macro$ calculates metrics for each label and find their unweighted mean. This does not take label imbalance into account.

\subsubsection{Baseline Approaches}
We introduce several deep-learning-based models as our baseline approaches without additional labeled data or external ontology resources. 

\begin{itemize}[leftmargin=*]



\item GRU is the basic Gated Recurrent Unit network.

\item MC-GRUs embeds the clinical feature via separate GRUs.

\item MC-GRUs$_t$ is pre-trained at the source dataset to obtain the parameters of corresponding GRUs.

\item ConCare (AAAI 2020) \cite{ma2020concare} embeds the feature sequences separately and uses the self-attention to model dynamic features and static baseline information.



\item TimeNet (IJCAI 2018) \cite{gupta2018using} maps variable-length clinical time series to fixed-dimensional feature vectors separately, and acts as an off-the-shelf feature extractor.
It is pre-trained on the UCR time series Repository.

\item \mname$_{stu}$ is the proposed \mname without knowledge distillation from the teacher model. 

\end{itemize}

\subsection{Experiment Results}

As is shown in Table~\ref{tab:results}\, \mname consistently outperforms both transfer-based and non-transfer-based baselines, demonstrating its ability to learn a robust representation. \mname achieves 6.8\% relative higher AUPRC and 1.5\% higher min(Se, P+) compared to the best state-of-the-art models ConCare and TimeNet.
Comparing all the methods with and without transfer mechanism, we can tell that utilizing preset knowledge from existing EMR can significantly promote the prediction performance of all models, indicating the effectiveness of the transfer learning mechanism. 
Moreover, we can see that \mname shows a higher performance than TimeNet. Both models employ feature-level transfer. However, our model \mname executes a more adaptive and reasonable transfer, which explains the superiority of our model's performance.
The knowledge distillation mechanism makes sense as well. Compared to the reduced $\mname_{stu}$ model, \mname also achieves higher performance in terms of all metrics. This indicates that developing knowledge distillation based on leveraging existing EMR can enhance healthcare prediction.



\subsection{Implications for COVID-19}

\begin{figure*}[]
\centering
\includegraphics[scale = 0.19]{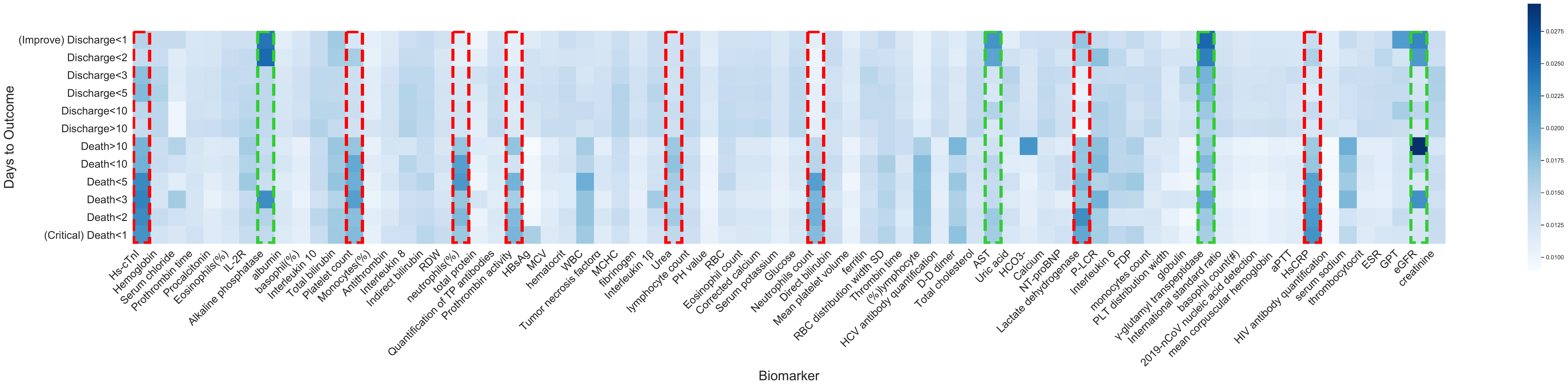}
\caption{Importance of Medical Features Differentiated by \mname. Features marked in red are \textit{fatal} indicators for patients in critical conditions, whose abnormal values indicate the emergency adverse outcome. Features marked in green are \textit{discharging} indicators, whose normal values indicate an overall improvement and soon discharging.}
\label{fig:attention_days}
\end{figure*}

To quantitatively identify the reasonability of feature recalibration from an overall perspective and extract useful medical knowledge,
we calculate the average importance weights of biomarkers for patients in diverse conditions.
As shown in Figure~\ref{fig:attention_days}, some essential medical knowledge learned by \mname can be summarized. 
Many of them have been proved or mentioned in COVID-19 related medical literature.
We also invite medical practitioners to evaluate the findings empirically.

\begin{itemize}[leftmargin=*]
    \item \textit{Hypersensitive Cardiac Troponin} has the most distinct difference between discharging and death cases, which means that \mname considers it as a significant mortality risk indicator. According to Chapman et al~\cite{chapman2020high}, troponin is elevated in one in five patients who have (confirmed) COVID-19 and that the presence of elevated troponin in COVID-19 may be associated with higher mortality risk. COVID-19 patients with elevated troponin may also be more likely to require ventilation and develop acute respiratory distress syndrome. For these patients, if clinicians are reluctant to measure cardiac troponin, they may ignore the plethora of ischaemic and non-ischaemic causes of myocardial injury related to COVID-19, which leads to a higher mortality rate.
    
    \item \textit{$\gamma$-Glutamyl Transpeptidase} (GGT) and \textit{Alkaline Phosphatase} are another two key features that \mname identified. They are significant biomarkers related to liver injury. According to a recent study published in the Lancet~\cite{zhang2020liver}, they are all related to COVID-19 patients' health status. Liver damage in mild cases of COVID-19 is often transient and can return to normal without any special treatment. However, when severe liver damage occurs, liver protective drugs have usually been given to such patients. In their case studies, GGT is elevated in 30 (54\%) of 56 patients with COVID-19 during hospitalization in their center. They also mention that elevated alkaline phosphatase levels are observed in one (1·8\%) of 56 patients with COVID-19 during hospitalization. 
    \item \textit{estimated Glomerular Filtration Rate} (eGFR), \textit{Urea} and \textit{Creatinie} are kidney injury-related biomarkers, and their difference between live and death cases are also distinct. According to Cheng et al~\cite{cheng2020kidney}, for on admission COVID-19 patients, \textit{Creatinie} and \textit{Urea} are elevated in 14.4\% and 13.1\% of the patients, respectively. eGFR < 60 ml/min per 1.73 $m^{2}$ is reported in 13.1\% of patients. Compared with patients with normal \textit{Creatinie}, those who entered the hospital with an elevated \textit{Creatinie} are older and more severely ill. The incidence of in-hospital death in patients with elevated baseline serum creatinine is 33.7\%, which is significantly higher than in those with normal baseline serum creatinine (13.2\%). 
\end{itemize}

\begin{table}[]
\small 
\centering
\caption{Mortality Prediction Performance for ESRD}
\label{tab:esrdresults}
\begin{tabular}{ccccc} 
\toprule
 &  & \multicolumn{3}{c}{Mortality Prediction Performance} \\
Methods & Trans. & AUPRC & AUROC & min(Se, P+)  \\ 
\midrule
 GRU & $\times$ & 0.7142\,(.0883) & 0.8094\,(.0547) & 0.6668\,(.0544)  \\
 MC-GRU & $\times$ & 0.7335\,(.0757) & 0.8193\,(.0468) & 0.6670\,(.0645) \\






ConCare & $\times$ & 0.7291\,(.0827) & 0.8259\,(.0456) & 0.6784\,(.0573)\\

TimeNet & $\surd$ & 0.6328\,(.0310) & 0.7311\,(.0262) & 0.5926\,(.0194) \\

\hline

MC-GRU$_{t}$ & $\surd$ & 0.7466\,(.0486) & 0.8260\,(.0330) & 0.6963\,(.0306) \\

\mname $_{stu}$ & $\surd$ & 0.7414\,(.0692) & 0.8263\,(.0427) & 0.6723\,(.0523)\\

\mname & $\surd$ & \textbf{0.7614}\,(.0584) & \textbf{0.8361}\,(.0385) & \textbf{0.7046}\,(.0353)\\

\bottomrule
\end{tabular}
\end{table}

\section{Extended Dataset Evaluation}

In order to further verify the generality of \mname, we also conduct an additional experiment on end-stage renal disease (ESRD) dataset.
We take the ESRD dataset as the target dataset and perform the mortality prediction.
Currently, many people are suffered from ESRD in the world~\cite{tangri2011determining,isakova2011fibroblast}.
They face severe life threats and need lifelong treatments with periodic visits to the hospitals for multifarious tests (e.g., blood routine examination).
The whole procedure needs a dynamic patient health risk prediction to help patients recover smoothly and prevent the adverse outcome, based on the medical records collected along with the visits.
The core task of \mname is to learn the health status representation of the patient and perform the healthcare prediction.

In this study, all ESRD patients who received therapy from January 1, 2006, to March 1, 2018, in a real-world hospital are included to form this dataset. 
During and after data collection and analysis, we did not identify individual participants as the patients' names, and they were replaced by patient ID.
This study was approved by the Medical Scientific Research Ethical Committee.
We drop the patients whose all entries of one feature are missing and select the observed features in more than 60\% of patients' records. 
For missing values, we fill the missing front cells with the data backward to prevent future information leakage. 
If the patient's backward record is missing, we impute it with the first front observed record of the patient.

The cleaned dataset consists of 662 patients and 13,108 visits. The statistics of the ESRD dataset are presented in Table~\ref{tab:esrdstat}.
Medical features recorded in the ESRD target dataset are listed in Table~\ref{tab:esrdfeature}.
The mortality prediction task on ESRD datasets is defined as a binary classification task of predicting the death of a patient in one year.

For the binary classification task, we assess performance using the area under the receiver operating characteristic curve (AUROC), area under the precision-recall curve (AUPRC), and the minimum of precision and sensitivity Min(Se, P+).
According to Table~\ref{tab:esrdresults}, \mname consistently outperforms other baseline approaches for all metrics.
The experiment results verify the applicability of our proposed framework.
\mname can not only predict LOS for new EID, but also perform mortality prediction for ESRD, which is the disease with limited EMR.

\section{conclusions}

In this paper, we propose a transfer learning-based prognosis solution, \mname, to perform the length of stay prediction for patients with COVID-19. 
In order to embed the medical features robustly, the model is trained to imitate the teacher model's medical embedding behavior via knowledge distillation.
The experimental results on the real-world COVID-19 dataset show that \mname consistently outperforms several competitive baseline methods. More importantly, \mname identifies several key indicators (e.g., \textit{hs-cTnI}, \textit{hs-CRP} and \textit{Platelet Counts}) for patients with critical conditions, and those abnormal values indicate potential emergent adverse outcomes.
\mname also reveals that normal values of \textit{$\gamma$-GT}, \textit{AP} and \textit{eGFR} indicate the overall improvement of health and a possible early-discharging.
The medical findings extracted by \mname are empirically validated and confirmed by human experts and medical literature. We believe the proposed model, \mname, will significantly benefit the intelligent prognosis for tackling future emerging infectious diseases such as COVID-19.


\bibliographystyle{ACM-Reference-Format}
\bibliography{sample-base}


\begin{thebibliography}{40}


\ifx \showCODEN    \undefined \def \showCODEN     #1{\unskip}     \fi
\ifx \showDOI      \undefined \def \showDOI       #1{#1}\fi
\ifx \showISBNx    \undefined \def \showISBNx     #1{\unskip}     \fi
\ifx \showISBNxiii \undefined \def \showISBNxiii  #1{\unskip}     \fi
\ifx \showISSN     \undefined \def \showISSN      #1{\unskip}     \fi
\ifx \showLCCN     \undefined \def \showLCCN      #1{\unskip}     \fi
\ifx \shownote     \undefined \def \shownote      #1{#1}          \fi
\ifx \showarticletitle \undefined \def \showarticletitle #1{#1}   \fi
\ifx \showURL      \undefined \def \showURL       {\relax}        \fi
\providecommand\bibfield[2]{#2}
\providecommand\bibinfo[2]{#2}
\providecommand\natexlab[1]{#1}
\providecommand\showeprint[2][]{arXiv:#2}

\bibitem[\protect\citeauthoryear{Baytas, Xiao, Zhang, Wang, Jain, and
  Zhou}{Baytas et~al\mbox{.}}{2017}]%
        {baytas2017patient}
\bibfield{author}{\bibinfo{person}{Inci~M Baytas}, \bibinfo{person}{Cao Xiao},
  \bibinfo{person}{Xi Zhang}, \bibinfo{person}{Fei Wang},
  \bibinfo{person}{Anil~K Jain}, {and} \bibinfo{person}{Jiayu Zhou}.}
  \bibinfo{year}{2017}\natexlab{}.
\newblock \showarticletitle{Patient subtyping via time-aware LSTM networks}. In
  \bibinfo{booktitle}{\emph{Proceedings of the 23rd ACM SIGKDD International
  Conference on Knowledge Discovery and Data Mining}}. ACM,
  \bibinfo{pages}{65--74}.
\newblock


\bibitem[\protect\citeauthoryear{Chapman, Bularga, and Mills}{Chapman
  et~al\mbox{.}}{2020}]%
        {chapman2020high}
\bibfield{author}{\bibinfo{person}{Andrew~R Chapman}, \bibinfo{person}{Anda
  Bularga}, {and} \bibinfo{person}{Nicholas~L Mills}.}
  \bibinfo{year}{2020}\natexlab{}.
\newblock \showarticletitle{High-sensitivity cardiac troponin can be an ally in
  the fight against COVID-19}.
\newblock \bibinfo{journal}{\emph{Circulation}} (\bibinfo{year}{2020}).
\newblock


\bibitem[\protect\citeauthoryear{Cheng, Luo, Wang, Zhang, Wang, Dong, Li, Yao,
  Ge, and Xu}{Cheng et~al\mbox{.}}{2020}]%
        {cheng2020kidney}
\bibfield{author}{\bibinfo{person}{Yichun Cheng}, \bibinfo{person}{Ran Luo},
  \bibinfo{person}{Kun Wang}, \bibinfo{person}{Meng Zhang},
  \bibinfo{person}{Zhixiang Wang}, \bibinfo{person}{Lei Dong},
  \bibinfo{person}{Junhua Li}, \bibinfo{person}{Ying Yao},
  \bibinfo{person}{Shuwang Ge}, {and} \bibinfo{person}{Gang Xu}.}
  \bibinfo{year}{2020}\natexlab{}.
\newblock \showarticletitle{Kidney disease is associated with in-hospital death
  of patients with COVID-19}.
\newblock \bibinfo{journal}{\emph{Kidney International}}
  (\bibinfo{year}{2020}).
\newblock


\bibitem[\protect\citeauthoryear{Choi, Bahadori, Schuetz, Stewart, and
  Sun}{Choi et~al\mbox{.}}{2015}]%
        {Choi2015Doctor}
\bibfield{author}{\bibinfo{person}{Edward Choi}, \bibinfo{person}{Mohammad~Taha
  Bahadori}, \bibinfo{person}{Andy Schuetz}, \bibinfo{person}{Walter~F
  Stewart}, {and} \bibinfo{person}{Jimeng Sun}.}
  \bibinfo{year}{2015}\natexlab{}.
\newblock \showarticletitle{Doctor AI: Predicting Clinical Events via Recurrent
  Neural Networks}.
\newblock  (\bibinfo{year}{2015}).
\newblock


\bibitem[\protect\citeauthoryear{Choi, Bahadori, Song, Stewart, and Sun}{Choi
  et~al\mbox{.}}{2017}]%
        {choi2017gram}
\bibfield{author}{\bibinfo{person}{Edward Choi}, \bibinfo{person}{Mohammad~Taha
  Bahadori}, \bibinfo{person}{Le Song}, \bibinfo{person}{Walter~F Stewart},
  {and} \bibinfo{person}{Jimeng Sun}.} \bibinfo{year}{2017}\natexlab{}.
\newblock \showarticletitle{GRAM: graph-based attention model for healthcare
  representation learning}. In \bibinfo{booktitle}{\emph{Proceedings of the
  23rd ACM SIGKDD International Conference on Knowledge Discovery and Data
  Mining}}. ACM, \bibinfo{pages}{787--795}.
\newblock


\bibitem[\protect\citeauthoryear{Choi, Bahadori, Sun, Kulas, Schuetz, and
  Stewart}{Choi et~al\mbox{.}}{2016}]%
        {choi2016retain}
\bibfield{author}{\bibinfo{person}{Edward Choi}, \bibinfo{person}{Mohammad~Taha
  Bahadori}, \bibinfo{person}{Jimeng Sun}, \bibinfo{person}{Joshua Kulas},
  \bibinfo{person}{Andy Schuetz}, {and} \bibinfo{person}{Walter Stewart}.}
  \bibinfo{year}{2016}\natexlab{}.
\newblock \showarticletitle{Retain: An interpretable predictive model for
  healthcare using reverse time attention mechanism}. In
  \bibinfo{booktitle}{\emph{Advances in Neural Information Processing
  Systems}}. \bibinfo{pages}{3504--3512}.
\newblock


\bibitem[\protect\citeauthoryear{Choi, Xiao, Stewart, and Sun}{Choi
  et~al\mbox{.}}{2018}]%
        {choi2018mime}
\bibfield{author}{\bibinfo{person}{Edward Choi}, \bibinfo{person}{Cao Xiao},
  \bibinfo{person}{Walter Stewart}, {and} \bibinfo{person}{Jimeng Sun}.}
  \bibinfo{year}{2018}\natexlab{}.
\newblock \showarticletitle{Mime: Multilevel medical embedding of electronic
  health records for predictive healthcare}. In
  \bibinfo{booktitle}{\emph{Advances in Neural Information Processing
  Systems}}. \bibinfo{pages}{4547--4557}.
\newblock


\bibitem[\protect\citeauthoryear{Emami, Javanmardi, Pirbonyeh, and
  Akbari}{Emami et~al\mbox{.}}{2020}]%
        {emami2020prevalence}
\bibfield{author}{\bibinfo{person}{Amir Emami}, \bibinfo{person}{Fatemeh
  Javanmardi}, \bibinfo{person}{Neda Pirbonyeh}, {and} \bibinfo{person}{Ali
  Akbari}.} \bibinfo{year}{2020}\natexlab{}.
\newblock \showarticletitle{Prevalence of underlying diseases in hospitalized
  patients with COVID-19: a systematic review and meta-analysis}.
\newblock \bibinfo{journal}{\emph{Archives of academic emergency medicine}}
  \bibinfo{volume}{8}, \bibinfo{number}{1} (\bibinfo{year}{2020}).
\newblock


\bibitem[\protect\citeauthoryear{Esteban, Staeck, Baier, Yang, and
  Tresp}{Esteban et~al\mbox{.}}{2016}]%
        {esteban2016predicting}
\bibfield{author}{\bibinfo{person}{Crist{\'o}bal Esteban},
  \bibinfo{person}{Oliver Staeck}, \bibinfo{person}{Stephan Baier},
  \bibinfo{person}{Yinchong Yang}, {and} \bibinfo{person}{Volker Tresp}.}
  \bibinfo{year}{2016}\natexlab{}.
\newblock \showarticletitle{Predicting clinical events by combining static and
  dynamic information using recurrent neural networks}. In
  \bibinfo{booktitle}{\emph{Healthcare Informatics (ICHI), 2016 IEEE
  International Conference on}}. Ieee, \bibinfo{pages}{93--101}.
\newblock


\bibitem[\protect\citeauthoryear{Fu, Wang, Yuan, Chen, Ao, Fitzpatrick, Li,
  Zhou, Lin, Duan, et~al\mbox{.}}{Fu et~al\mbox{.}}{2020}]%
        {fu2020clinical}
\bibfield{author}{\bibinfo{person}{Leiwen Fu}, \bibinfo{person}{Bingyi Wang},
  \bibinfo{person}{Tanwei Yuan}, \bibinfo{person}{Xiaoting Chen},
  \bibinfo{person}{Yunlong Ao}, \bibinfo{person}{Tom Fitzpatrick},
  \bibinfo{person}{Peiyang Li}, \bibinfo{person}{Yiguo Zhou},
  \bibinfo{person}{Yifan Lin}, \bibinfo{person}{Qibin Duan}, {et~al\mbox{.}}}
  \bibinfo{year}{2020}\natexlab{}.
\newblock \showarticletitle{Clinical characteristics of coronavirus disease
  2019 (COVID-19) in China: a systematic review and meta-analysis}.
\newblock \bibinfo{journal}{\emph{Journal of Infection}}
  (\bibinfo{year}{2020}).
\newblock


\bibitem[\protect\citeauthoryear{Gao, Xiao, Wang, Tang, Glass, and Sun}{Gao
  et~al\mbox{.}}{2020}]%
        {gao2020stagenet}
\bibfield{author}{\bibinfo{person}{Junyi Gao}, \bibinfo{person}{Cao Xiao},
  \bibinfo{person}{Yasha Wang}, \bibinfo{person}{Wen Tang},
  \bibinfo{person}{Lucas~M Glass}, {and} \bibinfo{person}{Jimeng Sun}.}
  \bibinfo{year}{2020}\natexlab{}.
\newblock \showarticletitle{StageNet: Stage-Aware Neural Networks for Health
  Risk Prediction}. In \bibinfo{booktitle}{\emph{Proceedings of The Web
  Conference 2020}}. \bibinfo{pages}{530--540}.
\newblock


\bibitem[\protect\citeauthoryear{Gupta, Malhotra, Vig, and Shroff}{Gupta
  et~al\mbox{.}}{2018a}]%
        {gupta2018transfer}
\bibfield{author}{\bibinfo{person}{Priyanka Gupta}, \bibinfo{person}{Pankaj
  Malhotra}, \bibinfo{person}{Lovekesh Vig}, {and} \bibinfo{person}{Gautam
  Shroff}.} \bibinfo{year}{2018}\natexlab{a}.
\newblock \showarticletitle{Transfer Learning for Clinical Time Series Analysis
  using Recurrent Neural Networks}.
\newblock \bibinfo{journal}{\emph{arXiv preprint arXiv:1807.01705}}
  (\bibinfo{year}{2018}).
\newblock


\bibitem[\protect\citeauthoryear{Gupta, Malhotra, Vig, and Shroff}{Gupta
  et~al\mbox{.}}{2018b}]%
        {gupta2018using}
\bibfield{author}{\bibinfo{person}{Priyanka Gupta}, \bibinfo{person}{Pankaj
  Malhotra}, \bibinfo{person}{Lovekesh Vig}, {and} \bibinfo{person}{Gautam
  Shroff}.} \bibinfo{year}{2018}\natexlab{b}.
\newblock \showarticletitle{Using Features from Pre-trained TimeNet for
  Clinical Predictions}. In \bibinfo{booktitle}{\emph{The 3rd International
  Workshop on Knowledge Discovery in Healthcare Data at IJCAI}}.
\newblock


\bibitem[\protect\citeauthoryear{Harutyunyan, Khachatrian, Kale, Ver~Steeg, and
  Galstyan}{Harutyunyan et~al\mbox{.}}{2019}]%
        {harutyunyan2019multitask}
\bibfield{author}{\bibinfo{person}{Hrayr Harutyunyan}, \bibinfo{person}{Hrant
  Khachatrian}, \bibinfo{person}{David~C Kale}, \bibinfo{person}{Greg
  Ver~Steeg}, {and} \bibinfo{person}{Aram Galstyan}.}
  \bibinfo{year}{2019}\natexlab{}.
\newblock \showarticletitle{Multitask learning and benchmarking with clinical
  time series data}.
\newblock \bibinfo{journal}{\emph{Scientific data}} \bibinfo{volume}{6},
  \bibinfo{number}{1} (\bibinfo{year}{2019}), \bibinfo{pages}{1--18}.
\newblock


\bibitem[\protect\citeauthoryear{Heo, Lee, Kim, Lee, Kim, Yang, and Hwang}{Heo
  et~al\mbox{.}}{2018}]%
        {heo2018uncertainty}
\bibfield{author}{\bibinfo{person}{Jay Heo}, \bibinfo{person}{Hae~Beom Lee},
  \bibinfo{person}{Saehoon Kim}, \bibinfo{person}{Juho Lee},
  \bibinfo{person}{Kwang~Joon Kim}, \bibinfo{person}{Eunho Yang}, {and}
  \bibinfo{person}{Sung~Ju Hwang}.} \bibinfo{year}{2018}\natexlab{}.
\newblock \showarticletitle{Uncertainty-aware attention for reliable
  interpretation and prediction}. In \bibinfo{booktitle}{\emph{Advances in
  Neural Information Processing Systems}}. \bibinfo{pages}{909--918}.
\newblock


\bibitem[\protect\citeauthoryear{Huang, Wang, Li, Ren, Zhao, Hu, Zhang, Fan,
  Xu, Gu, et~al\mbox{.}}{Huang et~al\mbox{.}}{2020}]%
        {huang2020clinical}
\bibfield{author}{\bibinfo{person}{Chaolin Huang}, \bibinfo{person}{Yeming
  Wang}, \bibinfo{person}{Xingwang Li}, \bibinfo{person}{Lili Ren},
  \bibinfo{person}{Jianping Zhao}, \bibinfo{person}{Yi Hu}, \bibinfo{person}{Li
  Zhang}, \bibinfo{person}{Guohui Fan}, \bibinfo{person}{Jiuyang Xu},
  \bibinfo{person}{Xiaoying Gu}, {et~al\mbox{.}}}
  \bibinfo{year}{2020}\natexlab{}.
\newblock \showarticletitle{Clinical features of patients infected with 2019
  novel coronavirus in Wuhan, China}.
\newblock \bibinfo{journal}{\emph{The lancet}} \bibinfo{volume}{395},
  \bibinfo{number}{10223} (\bibinfo{year}{2020}), \bibinfo{pages}{497--506}.
\newblock


\bibitem[\protect\citeauthoryear{Isakova, Xie, Yang, Xie, Anderson, Scialla,
  Wahl, Guti{\'e}rrez, Steigerwalt, He, et~al\mbox{.}}{Isakova
  et~al\mbox{.}}{2011}]%
        {isakova2011fibroblast}
\bibfield{author}{\bibinfo{person}{Tamara Isakova}, \bibinfo{person}{Huiliang
  Xie}, \bibinfo{person}{Wei Yang}, \bibinfo{person}{Dawei Xie},
  \bibinfo{person}{Amanda~Hyre Anderson}, \bibinfo{person}{Julia Scialla},
  \bibinfo{person}{Patricia Wahl}, \bibinfo{person}{Orlando~M Guti{\'e}rrez},
  \bibinfo{person}{Susan Steigerwalt}, \bibinfo{person}{Jiang He},
  {et~al\mbox{.}}} \bibinfo{year}{2011}\natexlab{}.
\newblock \showarticletitle{Fibroblast growth factor 23 and risks of mortality
  and end-stage renal disease in patients with chronic kidney disease}.
\newblock \bibinfo{journal}{\emph{Jama}} \bibinfo{volume}{305},
  \bibinfo{number}{23} (\bibinfo{year}{2011}), \bibinfo{pages}{2432--2439}.
\newblock


\bibitem[\protect\citeauthoryear{Ji, Ma, Peppelenbosch, and Pan}{Ji
  et~al\mbox{.}}{2020}]%
        {ji2020potential}
\bibfield{author}{\bibinfo{person}{Yunpeng Ji}, \bibinfo{person}{Zhongren Ma},
  \bibinfo{person}{Maikel~P Peppelenbosch}, {and} \bibinfo{person}{Qiuwei
  Pan}.} \bibinfo{year}{2020}\natexlab{}.
\newblock \showarticletitle{Potential association between COVID-19 mortality
  and health-care resource availability}.
\newblock \bibinfo{journal}{\emph{The Lancet Global Health}}
  \bibinfo{volume}{8}, \bibinfo{number}{4} (\bibinfo{year}{2020}),
  \bibinfo{pages}{e480}.
\newblock


\bibitem[\protect\citeauthoryear{Kingma and Ba}{Kingma and Ba}{2014}]%
        {kingma2014adam}
\bibfield{author}{\bibinfo{person}{Diederik~P Kingma} {and}
  \bibinfo{person}{Jimmy Ba}.} \bibinfo{year}{2014}\natexlab{}.
\newblock \showarticletitle{Adam: A method for stochastic optimization}.
\newblock \bibinfo{journal}{\emph{arXiv preprint arXiv:1412.6980}}
  (\bibinfo{year}{2014}).
\newblock


\bibitem[\protect\citeauthoryear{Lauer, Grantz, Bi, Jones, Zheng, Meredith,
  Azman, Reich, and Lessler}{Lauer et~al\mbox{.}}{2020}]%
        {lauer2020incubation}
\bibfield{author}{\bibinfo{person}{Stephen~A Lauer}, \bibinfo{person}{Kyra~H
  Grantz}, \bibinfo{person}{Qifang Bi}, \bibinfo{person}{Forrest~K Jones},
  \bibinfo{person}{Qulu Zheng}, \bibinfo{person}{Hannah~R Meredith},
  \bibinfo{person}{Andrew~S Azman}, \bibinfo{person}{Nicholas~G Reich}, {and}
  \bibinfo{person}{Justin Lessler}.} \bibinfo{year}{2020}\natexlab{}.
\newblock \showarticletitle{The incubation period of coronavirus disease 2019
  (COVID-19) from publicly reported confirmed cases: estimation and
  application}.
\newblock \bibinfo{journal}{\emph{Annals of internal medicine}}
  \bibinfo{volume}{172}, \bibinfo{number}{9} (\bibinfo{year}{2020}),
  \bibinfo{pages}{577--582}.
\newblock


\bibitem[\protect\citeauthoryear{Lee, Luo, Ngiam, Zhang, Zheng, Chen, Ooi, and
  Yip}{Lee et~al\mbox{.}}{2017}]%
        {lee2017big}
\bibfield{author}{\bibinfo{person}{Chonho Lee}, \bibinfo{person}{Zhaojing Luo},
  \bibinfo{person}{Kee~Yuan Ngiam}, \bibinfo{person}{Meihui Zhang},
  \bibinfo{person}{Kaiping Zheng}, \bibinfo{person}{Gang Chen},
  \bibinfo{person}{Beng~Chin Ooi}, {and} \bibinfo{person}{Wei Luen~James Yip}.}
  \bibinfo{year}{2017}\natexlab{}.
\newblock \showarticletitle{Big healthcare data analytics: Challenges and
  applications}.
\newblock In \bibinfo{booktitle}{\emph{Handbook of Large-Scale Distributed
  Computing in Smart Healthcare}}. \bibinfo{publisher}{Springer},
  \bibinfo{pages}{11--41}.
\newblock


\bibitem[\protect\citeauthoryear{Lee, Park, Joo, and Moon}{Lee
  et~al\mbox{.}}{2018}]%
        {lee2018diagnosis}
\bibfield{author}{\bibinfo{person}{Wonsung Lee}, \bibinfo{person}{Sungrae
  Park}, \bibinfo{person}{Weonyoung Joo}, {and} \bibinfo{person}{Il-Chul
  Moon}.} \bibinfo{year}{2018}\natexlab{}.
\newblock \showarticletitle{Diagnosis Prediction via Medical Context Attention
  Networks Using Deep Generative Modeling}. In \bibinfo{booktitle}{\emph{2018
  IEEE International Conference on Data Mining (ICDM)}}. IEEE,
  \bibinfo{pages}{1104--1109}.
\newblock


\bibitem[\protect\citeauthoryear{Ma, You, Xiao, Chitta, Zhou, and Gao}{Ma
  et~al\mbox{.}}{2018b}]%
        {ma2018kame}
\bibfield{author}{\bibinfo{person}{Fenglong Ma}, \bibinfo{person}{Quanzeng
  You}, \bibinfo{person}{Houping Xiao}, \bibinfo{person}{Radha Chitta},
  \bibinfo{person}{Jing Zhou}, {and} \bibinfo{person}{Jing Gao}.}
  \bibinfo{year}{2018}\natexlab{b}.
\newblock \showarticletitle{Kame: Knowledge-based attention model for diagnosis
  prediction in healthcare}. In \bibinfo{booktitle}{\emph{Proceedings of the
  27th ACM International Conference on Information and Knowledge Management}}.
  ACM, \bibinfo{pages}{743--752}.
\newblock


\bibitem[\protect\citeauthoryear{Ma, Gao, Wang, Zhang, Wang, Ruan, Tang, Gao,
  and Ma}{Ma et~al\mbox{.}}{2020a}]%
        {ma2020adacare}
\bibfield{author}{\bibinfo{person}{Liantao Ma}, \bibinfo{person}{Junyi Gao},
  \bibinfo{person}{Yasha Wang}, \bibinfo{person}{Chaohe Zhang},
  \bibinfo{person}{Jiangtao Wang}, \bibinfo{person}{Wenjie Ruan},
  \bibinfo{person}{Wen Tang}, \bibinfo{person}{Xin Gao}, {and}
  \bibinfo{person}{Xinyu Ma}.} \bibinfo{year}{2020}\natexlab{a}.
\newblock \showarticletitle{AdaCare: AdaCare: Explainable Clinical Health
  Status Representation Learning via Scale-Adaptive Feature Extraction and
  Recalibration}. In \bibinfo{booktitle}{\emph{Thirty-Fourth AAAI Conference on
  Artificial Intelligence}}.
\newblock


\bibitem[\protect\citeauthoryear{Ma, Zhang, Wang, Ruan, Wang, Tang, Ma, Gao,
  and Gao}{Ma et~al\mbox{.}}{2020b}]%
        {ma2020concare}
\bibfield{author}{\bibinfo{person}{Liantao Ma}, \bibinfo{person}{Chaohe Zhang},
  \bibinfo{person}{Yasha Wang}, \bibinfo{person}{Wenjie Ruan},
  \bibinfo{person}{Jiangtao Wang}, \bibinfo{person}{Wen Tang},
  \bibinfo{person}{Xinyu Ma}, \bibinfo{person}{Xin Gao}, {and}
  \bibinfo{person}{Junyi Gao}.} \bibinfo{year}{2020}\natexlab{b}.
\newblock \showarticletitle{ConCare: Personalized Clinical Feature Embedding
  via Capturing the Healthcare Context}. In
  \bibinfo{booktitle}{\emph{Thirty-Fourth AAAI Conference on Artificial
  Intelligence}}.
\newblock


\bibitem[\protect\citeauthoryear{Ma, Xiao, and Wang}{Ma et~al\mbox{.}}{2018a}]%
        {ma2018health}
\bibfield{author}{\bibinfo{person}{Tengfei Ma}, \bibinfo{person}{Cao Xiao},
  {and} \bibinfo{person}{Fei Wang}.} \bibinfo{year}{2018}\natexlab{a}.
\newblock \showarticletitle{Health-ATM: A Deep Architecture for Multifaceted
  Patient Health Record Representation and Risk Prediction}. In
  \bibinfo{booktitle}{\emph{Proceedings of the 2018 SIAM International
  Conference on Data Mining}}. SIAM, \bibinfo{pages}{261--269}.
\newblock


\bibitem[\protect\citeauthoryear{of~Health, Services, et~al\mbox{.}}{of~Health
  et~al\mbox{.}}{2017}]%
        {us2017pandemic}
\bibfield{author}{\bibinfo{person}{US~Department of Health},
  \bibinfo{person}{Human Services}, {et~al\mbox{.}}}
  \bibinfo{year}{2017}\natexlab{}.
\newblock \showarticletitle{Pandemic influenza plan: 2017 Update}.
\newblock \bibinfo{journal}{\emph{URL https://www. cdc.
  gov/flu/pandemic-resources/pdf/pan-flu-report-2017v2. pdf}}
  (\bibinfo{year}{2017}).
\newblock


\bibitem[\protect\citeauthoryear{Pedersen, Eliasen, and Henriksen}{Pedersen
  et~al\mbox{.}}{1990}]%
        {pedersen1990prospective}
\bibfield{author}{\bibinfo{person}{T Pedersen}, \bibinfo{person}{K Eliasen},
  {and} \bibinfo{person}{Eet~al Henriksen}.} \bibinfo{year}{1990}\natexlab{}.
\newblock \showarticletitle{A prospective study of mortality associated with
  anaesthesia and surgery: risk indicators of mortality in hospital}.
\newblock \bibinfo{journal}{\emph{Acta Anaesthesiologica Scandinavica}}
  \bibinfo{volume}{34}, \bibinfo{number}{3} (\bibinfo{year}{1990}),
  \bibinfo{pages}{176--182}.
\newblock


\bibitem[\protect\citeauthoryear{Pham, Tran, Phung, and Venkatesh}{Pham
  et~al\mbox{.}}{2016}]%
        {pham2016deepcare}
\bibfield{author}{\bibinfo{person}{Trang Pham}, \bibinfo{person}{Truyen Tran},
  \bibinfo{person}{Dinh Phung}, {and} \bibinfo{person}{Svetha Venkatesh}.}
  \bibinfo{year}{2016}\natexlab{}.
\newblock \showarticletitle{Deepcare: A deep dynamic memory model for
  predictive medicine}. In \bibinfo{booktitle}{\emph{Pacific-Asia Conference on
  Knowledge Discovery and Data Mining}}. Springer, \bibinfo{pages}{30--41}.
\newblock


\bibitem[\protect\citeauthoryear{Purushotham, Meng, Che, and Liu}{Purushotham
  et~al\mbox{.}}{2017}]%
        {purushotham2017benchmark}
\bibfield{author}{\bibinfo{person}{Sanjay Purushotham},
  \bibinfo{person}{Chuizheng Meng}, \bibinfo{person}{Zhengping Che}, {and}
  \bibinfo{person}{Yan Liu}.} \bibinfo{year}{2017}\natexlab{}.
\newblock \showarticletitle{Benchmark of Deep Learning Models on Large
  Healthcare MIMIC Datasets.}
\newblock \bibinfo{journal}{\emph{arXiv: Learning}} (\bibinfo{year}{2017}).
\newblock


\bibitem[\protect\citeauthoryear{Qiao, Zhao, Xiao, Li, Qin, and Wang}{Qiao
  et~al\mbox{.}}{2018}]%
        {qiao2018pairwise}
\bibfield{author}{\bibinfo{person}{Zhi Qiao}, \bibinfo{person}{Shiwan Zhao},
  \bibinfo{person}{Cao Xiao}, \bibinfo{person}{Xiang Li}, \bibinfo{person}{Yong
  Qin}, {and} \bibinfo{person}{Fei Wang}.} \bibinfo{year}{2018}\natexlab{}.
\newblock \showarticletitle{Pairwise-Ranking based Collaborative Recurrent
  Neural Networks for Clinical Event Prediction.}. In
  \bibinfo{booktitle}{\emph{IJCAI}}. \bibinfo{pages}{3520--3526}.
\newblock


\bibitem[\protect\citeauthoryear{Reddy and Aggarwal}{Reddy and
  Aggarwal}{2015}]%
        {reddy2015healthcare}
\bibfield{author}{\bibinfo{person}{Chandan~K Reddy} {and}
  \bibinfo{person}{Charu~C Aggarwal}.} \bibinfo{year}{2015}\natexlab{}.
\newblock \bibinfo{booktitle}{\emph{Healthcare data analytics}}.
\newblock \bibinfo{publisher}{Chapman and Hall/CRC}.
\newblock


\bibitem[\protect\citeauthoryear{Reyna, Josef, Jeter, Shashikumar, Westover,
  Nemati, Clifford, and Sharma}{Reyna et~al\mbox{.}}{2019}]%
        {reyna2019early}
\bibfield{author}{\bibinfo{person}{Matthew~A Reyna},
  \bibinfo{person}{Christopher~S Josef}, \bibinfo{person}{Russell Jeter},
  \bibinfo{person}{Supreeth~P Shashikumar}, \bibinfo{person}{M~Brandon
  Westover}, \bibinfo{person}{Shamim Nemati}, \bibinfo{person}{Gari~D
  Clifford}, {and} \bibinfo{person}{Ashish Sharma}.}
  \bibinfo{year}{2019}\natexlab{}.
\newblock \showarticletitle{Early prediction of sepsis from clinical data: the
  PhysioNet/Computing in Cardiology Challenge 2019}.
\newblock \bibinfo{journal}{\emph{Critical Care Medicine}}
  (\bibinfo{year}{2019}).
\newblock


\bibitem[\protect\citeauthoryear{Suresh, Gong, and Guttag}{Suresh
  et~al\mbox{.}}{2018}]%
        {suresh2018learning}
\bibfield{author}{\bibinfo{person}{Harini Suresh}, \bibinfo{person}{Jen~J
  Gong}, {and} \bibinfo{person}{John Guttag}.} \bibinfo{year}{2018}\natexlab{}.
\newblock \showarticletitle{Learning Tasks for Multitask Learning: Heterogenous
  Patient Populations in the ICU}.
\newblock \bibinfo{journal}{\emph{arXiv preprint arXiv:1806.02878}}
  (\bibinfo{year}{2018}).
\newblock


\bibitem[\protect\citeauthoryear{Tangri, Ansell, and Naimark}{Tangri
  et~al\mbox{.}}{2011}]%
        {tangri2011determining}
\bibfield{author}{\bibinfo{person}{Navdeep Tangri}, \bibinfo{person}{David
  Ansell}, {and} \bibinfo{person}{David Naimark}.}
  \bibinfo{year}{2011}\natexlab{}.
\newblock \showarticletitle{Determining factors that predict technique survival
  on peritoneal dialysis: application of regression and artificial neural
  network methods}.
\newblock \bibinfo{journal}{\emph{Nephron Clinical Practice}}
  \bibinfo{volume}{118}, \bibinfo{number}{2} (\bibinfo{year}{2011}),
  \bibinfo{pages}{c93--c100}.
\newblock


\bibitem[\protect\citeauthoryear{Wang, Fang, Cai, Wu, Gao, Min, Wang,
  et~al\mbox{.}}{Wang et~al\mbox{.}}{2020}]%
        {wang2020comorbid}
\bibfield{author}{\bibinfo{person}{Xinhui Wang}, \bibinfo{person}{Xuexian
  Fang}, \bibinfo{person}{Zhaoxian Cai}, \bibinfo{person}{Xiaotian Wu},
  \bibinfo{person}{Xiaotong Gao}, \bibinfo{person}{Junxia Min},
  \bibinfo{person}{Fudi Wang}, {et~al\mbox{.}}}
  \bibinfo{year}{2020}\natexlab{}.
\newblock \showarticletitle{Comorbid Chronic Diseases and Acute Organ Injuries
  Are Strongly Correlated with Disease Severity and Mortality among COVID-19
  Patients: A Systemic Review and Meta-Analysis}.
\newblock \bibinfo{journal}{\emph{Research}}  \bibinfo{volume}{2020}
  (\bibinfo{year}{2020}), \bibinfo{pages}{2402961}.
\newblock


\bibitem[\protect\citeauthoryear{Weissman, Crane-Droesch, Chivers, Luong,
  Hanish, Levy, Lubken, Becker, Draugelis, Anesi, et~al\mbox{.}}{Weissman
  et~al\mbox{.}}{2020}]%
        {weissman2020locally}
\bibfield{author}{\bibinfo{person}{Gary~E Weissman}, \bibinfo{person}{Andrew
  Crane-Droesch}, \bibinfo{person}{Corey Chivers}, \bibinfo{person}{ThaiBinh
  Luong}, \bibinfo{person}{Asaf Hanish}, \bibinfo{person}{Michael~Z Levy},
  \bibinfo{person}{Jason Lubken}, \bibinfo{person}{Michael Becker},
  \bibinfo{person}{Michael~E Draugelis}, \bibinfo{person}{George~L Anesi},
  {et~al\mbox{.}}} \bibinfo{year}{2020}\natexlab{}.
\newblock \showarticletitle{Locally informed simulation to predict hospital
  capacity needs during the COVID-19 pandemic}.
\newblock \bibinfo{journal}{\emph{Annals of internal medicine}}
  (\bibinfo{year}{2020}).
\newblock


\bibitem[\protect\citeauthoryear{Yan, Zhang, Goncalves, Xiao, Wang, Guo, Sun,
  Tang, Jing, Zhang, et~al\mbox{.}}{Yan et~al\mbox{.}}{2020}]%
        {yan2020interpretable}
\bibfield{author}{\bibinfo{person}{Li Yan}, \bibinfo{person}{Hai-Tao Zhang},
  \bibinfo{person}{Jorge Goncalves}, \bibinfo{person}{Yang Xiao},
  \bibinfo{person}{Maolin Wang}, \bibinfo{person}{Yuqi Guo},
  \bibinfo{person}{Chuan Sun}, \bibinfo{person}{Xiuchuan Tang},
  \bibinfo{person}{Liang Jing}, \bibinfo{person}{Mingyang Zhang},
  {et~al\mbox{.}}} \bibinfo{year}{2020}\natexlab{}.
\newblock \showarticletitle{An interpretable mortality prediction model for
  COVID-19 patients}.
\newblock \bibinfo{journal}{\emph{Nature Machine Intelligence}}
  (\bibinfo{year}{2020}), \bibinfo{pages}{1--6}.
\newblock


\bibitem[\protect\citeauthoryear{Zhang, Shi, and Wang}{Zhang
  et~al\mbox{.}}{2020}]%
        {zhang2020liver}
\bibfield{author}{\bibinfo{person}{Chao Zhang}, \bibinfo{person}{Lei Shi},
  {and} \bibinfo{person}{Fu-Sheng Wang}.} \bibinfo{year}{2020}\natexlab{}.
\newblock \showarticletitle{Liver injury in COVID-19: management and
  challenges}.
\newblock \bibinfo{journal}{\emph{The lancet Gastroenterology \& hepatology}}
  \bibinfo{volume}{5}, \bibinfo{number}{5} (\bibinfo{year}{2020}),
  \bibinfo{pages}{428--430}.
\newblock


\bibitem[\protect\citeauthoryear{Zheng, Gao, Ngiam, Ooi, and Yip}{Zheng
  et~al\mbox{.}}{2017}]%
        {zheng2017resolving}
\bibfield{author}{\bibinfo{person}{Kaiping Zheng}, \bibinfo{person}{Jinyang
  Gao}, \bibinfo{person}{Kee~Yuan Ngiam}, \bibinfo{person}{Beng~Chin Ooi},
  {and} \bibinfo{person}{Wei Luen~James Yip}.} \bibinfo{year}{2017}\natexlab{}.
\newblock \showarticletitle{Resolving the bias in electronic medical records}.
  In \bibinfo{booktitle}{\emph{Proceedings of the 23rd ACM SIGKDD International
  Conference on Knowledge Discovery and Data Mining}}. ACM,
  \bibinfo{pages}{2171--2180}.
\newblock


\end{thebibliography}

\appendix

\section{Appendix}

\subsection{Experiment Environment}
The experiment environment is a machine equipped with CPU: Intel Xeon E5-2630, 256GB RAM, and GPU: Nvidia RTX8000. The code is implemented based on Pytorch 1.5.0. 
To train the model, we use Adam~\cite{kingma2014adam} with the batch size of 256, and the learning rate is set to $1e-3$. To fairly compare different approaches, the hyper-parameters of the baseline models are fine-tuned by the grid-searching strategy.

\begin{table}[]
  \caption{Statistics of Length of Stay for COVID-19}
  \label{tab:statlos}
  \begin{tabular}{lccc}
    \toprule
     &All& Discharging &Death\\
    \midrule
    Avg. \# of admissions per patient & 16.29 & 16 & 16.7\\
    Avg. \# of LOS per patient & 10.85 & 13.45 & 7.85\\
    Avg. \# of LOS per admission & 8.2 & 9.71 & 6.52 \\
  \bottomrule
\end{tabular}
\end{table}

\begin{figure}[h]
\centering
\includegraphics[scale = 0.45]{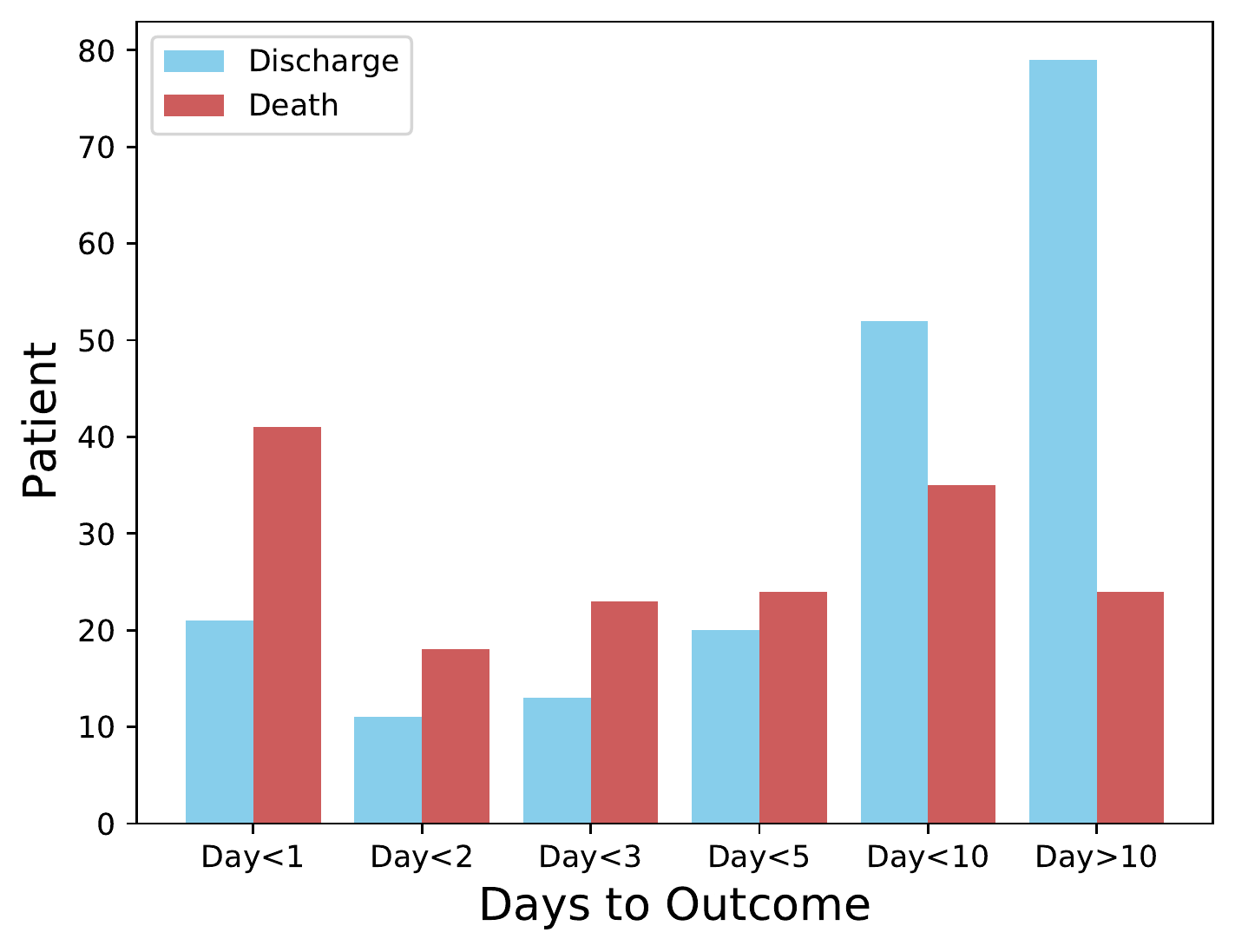}
\caption{Days to Outcome for Admissions of Patients with COVID-19}
\label{fig:days2outcome}
\end{figure}

\begin{table}[]
  \caption{Features Recorded in COVID-19 and PhysioNet Dataset}
  \label{tab:feature}
  \begin{tabular}{ccc}
  \hline
  
  \hline
  
  Shared Features & Private in PhysioNet & Private in COVID-19 \\
  \hline
    Hs-cTnI & Heart rate & $\gamma$-GT\\
    Hemoglobin & Pulse oximetry & Procalcitonin\\
    Serum chloride & Temperature & Albumin\\
    Alkaline phosphatase & Systolic BP& HBsAg\\
    Total bilirubin & MAP & Globulin\\
    Creatinine & Platelet count & HsCRP\\
    Hematocrit & Diastolic BP & Serum sodium\\
    WBC & Respiration rate& Red blood cell count\\
    Fibrinogen & EtCO2& nucleic acid detection\\
    Urea & Excess HCO3& Monocytes\\
    PH value& FiO2& Antithrombin\\
    Serum potassium& PaCO2& Total protein\\
    Glucose & SaO2& HCV-AQ\\
    Direct bilirubin& AST& Total cholesterol\\
    HCO3- & Lactic acid & Lactate dehydrogenase\\
    Calcium & Magnesium& HIV-AQ\\
    aPTT & Phosphate&...\\
  \bottomrule
\end{tabular}
\end{table}

\begin{table}
\centering
  \caption{Statistics of ESRD Dataset}
  \label{tab:esrdstat}
  \begin{tabular}{lc}

     \hline
    Statistic&Value\\
     \hline

    \# patients & 656\\
    \# patient with diabetes & 244\\
    \# patient died & 261\\
    \# visit & 13091\\
     \# visit in high risk & 1196\\
      \# visit in low risk & 10804\\
      \# visit in uncertain & 1091\\
    \% female & 49\% \\
     \hline
\end{tabular}
\end{table}

\begin{table}[]
  \caption{Features Recorded in ESRD Target Dataset}
  \label{tab:esrdfeature}
  \begin{tabular}{cc}
  \hline
  
  \hline
  
  Shared Features & Private in ESRD \\
  \hline
    Systolic BP & Sodium \\
    Diastolic BP& CO2CP\\
    Urea & Albumin \\
    Calcium & hs-CRP\\
    Chloride & Weight\\
    Creatinine & Amount\\
    Glucose & \\
    Phosphate & \\
    Potassium & \\
    Hemoglobin & \\
    WBC Count& \\
  \bottomrule
\end{tabular}
\end{table}

\begin{algorithm}
\caption{\mname $(\mathcal{R}_{src}, \mathcal{R}_{tar})$}
\label{alg:full_model}
\begin{algorithmic}[1]
\State Randomly initializing parameters in Teacher Model \mname $_{tea}$
\While {not convergence}:
    \State Compute $\hat{y}_{T, src}, s_{tea} =  \mname _{tea}(\mathcal{R}_{src})$
    \State Compute $\mathcal{L}_{pred} = \mathrm{CE}(\hat{y}_{T,src}, y_{T,src})$
    \State Compute $\mathcal{L}_{tea} = \mathcal{L}_{pred}$
    \State Update parameters of $\mname _{tea}$ by optimizing $\mathcal{L}_{tea}$ using back-propagation
\EndWhile

\State Randomly initializing parameters in Student Model \mname $_{stu}$
\While {not convergence}:
    \State Compute $\hat{y}_{T, src},  \hat{s}_{tea} =  \mname _{stu}(\tilde{\mathcal{R}}_{src})$
    \State Compute $\mathcal{L}_{pred} = \mathrm{CE}(\hat{y}_{T,src}, y_{T,src})$, $ \quad \mathcal{L}_{emb} = D_{KL} (\mathrm{Softmax}(\hat{s}_{tea}) || \mathrm{Softmax}(s_{tea}) )$
    \State Compute $\mathcal{L}_{stu} = \mathcal{L}_{pred} + \mathcal{L}_{emb}$
    \State Update parameters of $\mname _{stu}$ by optimizing $\mathcal{L}_{stu}$ using back-propagation
\EndWhile

\State Transfer parameters of \textbf{shared GRUs} from \mname $_{stu}$ to Target Model \mname $_{tar}$, and randomly initializing other parameters in \mname $_{tar}$
\While {not convergence}:
    \State Compute $\hat{y}_{T, tar} =  \mname _{stu}(\mathcal{R}_{tar})$
    \State Compute $\mathcal{L}_{pred} = \mathrm{CE}(\hat{y}_{T,tar}, y_{T,tar})$
    \State Compute $\mathcal{L}_{tar} = \mathcal{L}_{pred}$
    \State Update parameters of $\mname _{tar}$ by optimizing $\mathcal{L}_{tar}$ using back-propagation
\EndWhile
\end{algorithmic}
\end{algorithm}



    

\end{document}